\documentclass[11pt,a4paper]{article}
\usepackage[hyperref]{acl2018}
\usepackage{times}
\usepackage{latexsym}
\usepackage{url}
\usepackage{svg}
\usepackage{latexsym}
\usepackage{graphicx} 
\usepackage{enumitem}
\usepackage{multirow}

\usepackage{colortbl}

\usepackage{url}

\aclfinalcopy 

\title{Recurrent Stacking of Layers for Compact Neural Machine Translation Models}

\author{Raj Dabre \and Atsushi Fujita \\
  National Institute of Information and Communications Technology \\
  3-5 Hikaridai, Seika-cho, Soraku-gun, Kyoto 619-0289, Japan \\
  {\tt \{raj.dabre,atsushi.fujita\}@nict.go.jp}\\}

\date{}

\begin{document}
\maketitle
\begin{abstract}
In neural machine translation (NMT), the most common practice is to stack a number of recurrent or feed-forward layers in the encoder and the decoder. As a result, the addition of each new layer improves the translation quality significantly. However, this also leads to a significant increase in the number of parameters. In this paper, we propose to share parameters across all the layers thereby leading to a recurrently stacked NMT model. We empirically show that the translation quality of a model that recurrently stacks a single layer 6 times is comparable to the translation quality of a model that stacks 6 separate layers. We also show that using pseudo-parallel corpora by back-translation leads to further significant improvements in translation quality. 
\end{abstract}

\section{Introduction}
Neural machine translation (NMT) 
\cite{DBLP:journals/corr/ChoMGBSB14,DBLP:journals/corr/SutskeverVL14,DBLP:journals/corr/BahdanauCB14:original} allows for end-to-end training of a translation system without needing to deal with word alignments, translation rules and complicated decoding algorithms, which are integral to statistical machine translation (SMT) \cite{koehn-EtAl:2007:PosterDemo}. In encoder-decoder based NMT models, one of the most commonly followed practices is the stacking of multiple recurrent\footnote{Recurrent across time-steps.}, convolutional or self-attentional feed-forward layers in the encoders and decoders with each layer having its own parameters. It has been empirically shown that such stacking leads to an improvement in translation quality, especially in resource rich resource scenarios. However, it also increases the size of the model by a significant amount.

In this paper, we propose to reduce the number of model parameters by sharing parameters across layers. In other words, our Recurrently Stacked NMT (RSNMT) model has the same size of a single layer NMT model. We evaluate our method on several publicly available data-sets and show that a RSNMT model with 6 recurrence steps gives results that are comparable to a 6-layer NMT model which does not use any recurrences. The contributions of this paper are as follows:
\begin{itemize}
    \item We propose a novel modification to the NMT architecture where parameters are shared across layers which we call Recurrently Stacked NMT or RSNMT.
    \item We use the {\it Transformer} \cite{NIPS2017_7181} architecture but our method is architecture independent.
    \item We experiment with several publicly available data-sets and empirically show the effectiveness of our approach. The language directions we experimented with are: Turkish-English and English-Turkish (WMT), Japanese-English (ALT, KFTT, GCP) and English-Japanese (GCP).
    \item We also experimented with using back-translated corpora and show that our method further benefits from the additional data.
    \item To the best of our knowledge, this is the first work that shows that it is possible to reduce the NMT model size by sharing parameters across layers and yet achieve results that are comparable to a model that does not share parameters across layers.
\end{itemize}

\section{Related Work}
The most prominent way of reducing the size of a neural model is knowledge distillation \cite{DBLP:journals/corr/HintonVD15}, which requires training a parent model which can be a time-consuming task. The work on zero-shot NMT \cite{gnmt16multi:original} shows that it is possible for multiple language pairs to share a single encoder and decoder without an appreciable loss in translation quality. However, this work does not consider sharing the parameters across the stacked layers in the encoder or the decoder. The work on Universal Transformer \cite{univtrans} shows that feeding the output of the multi-layer encoder (and decoder) to itself repeatedly leads to an improvement in quality for English-German translation. Our method is similar to this, except that our RSNMT model has the same size as that of a 1-layer NMT model and yet manages to approach the translation quality given by a 6-layer NMT model. We additionally show that the recurrent stacking of layers can benefit from back-translated data.

\begin{figure}[t]
    \begin{center}
          \includegraphics[height=9cm, width=8cm]{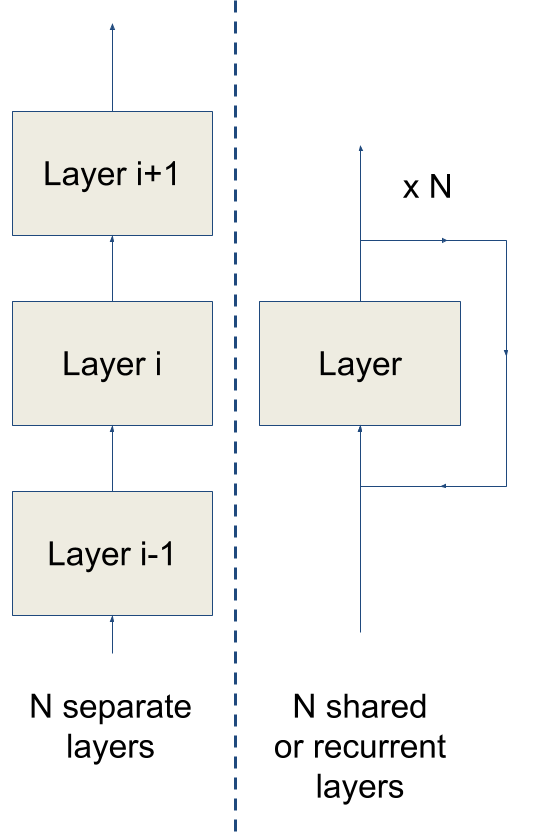}
     \end{center}
    \centering
  \vspace{-4mm}
      \caption{Vanilla NMT layer stacking (left) vs Recurrent NMT layer stacking (right)}
        \label{rsnmt}
    \vspace{-4mm}
\end{figure}

\section{Recurrent Stacked NMT}
\label{sec:overview}
Figure~\ref{rsnmt} illustrates our approach. The left-hand side shows the vanilla stacking of N layers where each layer in the neural network has its own parameters. The right-hand side shows our approach of stacking N layers where we use the same parameters for all layers. As a result of this sharing, the resultant neural model is technically a single layer model in which the same layer is recurrently stacked N times. This leads to a massive reduction in the size of the model.

\section{Experimental Settings}
\label{sec:settings}

\subsection{Data-sets and Languages}
We experimented with Japanese-English translation using the Asian language treebank (ALT) parallel corpus\footnote{http://www2.nict.go.jp/astrec-att/member/mutiyama/ALT/index.html} 
\cite{KYAWTHU16.435} and the Kyoto free translation task (KFTT) corpus\footnote{http://www.phontron.com/kftt} 
\cite{neubig11kftt}, both of which are publicly available. The ALT-JE task contains 18088, 1000, and 1018 sentences for training, development, and testing, respectively. The KFTT-JE task contains 440288, 1166, and 1160 sentences for training, development, and testing, respectively. We also experimented with Turkish-English and English-Turkish translation using the WMT 2018 corpus\footnote{http://www.statmt.org/wmt18/translation-task.html} which contains 207678, 3007 and 3010 sentences for training, development, and testing, respectively. Finally, we experimented with Japanese-English and English-Japanese translation using an in-house parallel corpus called the GCP corpus \cite{IMAMURA18.104,W18-2707}, which consists of 400000, 2000, 2000 sentences for training, development and testing, respectively. In addition, there are 1552475 lines of monolingual corpora for both Japanese and English, which we use for back-translation experiments. 

We tokenized the Japanese sentences in the KFTT and ALT corpora using the JUMAN\footnote{http://nlp.ist.i.kyoto-u.ac.jp/EN/index.php?JUMAN} \cite{kurohashi--EtAl:1994} morphological analyzer. We tokenized and lowercased the English sentences in KFTT and ALT using the {\it tokenizer.perl} and {\it lowercase.perl} scripts in Moses\footnote{http://www.statmt.org/moses}. The GCP corpora were available to us in a pre-tokenized and lowercased form. We did not tokenize the Turkish-English data in any way\footnote{tensor2tensor has an internal tokenization mechanism which was used for this language pair.}.

\subsection{NMT models} \label{sec:nmtmodels}
We trained and evaluated the following NMT models:
\begin{itemize}
    \item 6-layer model without any shared parameters across layers
    \item 1, 2, 3, 4, 5 and 6-layer models with parameters shared across all layers. These are referred to as 1, 2, 3, 4, 5 and 6 recurrently stacked NMT models
\end{itemize}

\subsection{Implementation and Model Settings}
We used the open-source implementation of the Transformer model \cite{NIPS2017_7181} in {\it tensor2tensor}\footnote{https://github.com/tensorflow/tensor2tensor} for all our NMT experiments. We implemented our approach using the version 1.6 branch of tensor2tensor. We used the Transformer because it is the current state-of-the-art NMT model. However, our approach of sharing parameters across layers is implementation and model independent. For training, we used the default model settings corresponding to {\it transformer\_base\_single\_gpu} in the implementation and to {\it base\_model} in \cite{NIPS2017_7181} with the exception of the number of sub-words, training iterations and number of GPUs. These numbers vary as we train the models to convergence. We used the tensor2tensor internal sub-word segmenter for simplicity. For the GCP corpora we used separate 16000 sub-word vocabularies and trained all models on 1 GPU with 60000 iterations for English-Japanese and 120000 iterations for Japanese-English. For the KFTT corpus we used separate 16000 sub-word vocabularies and trained all models on 1 GPU for 160000 iterations. For the ALT corpus we used separate 8000 sub-word vocabularies and trained all models on 1 GPU for 40000 iterations. For the WMT corpus we used a joint\footnote{We did this to exploit cognates across Turkish and English.} 16000 sub-word vocabulary and trained all models on 4 GPUs for 50000 iterations.

We averaged the last 10 checkpoints and decoded the test set sentences with a beam size of 4 and length penalty  of $\alpha=0.6$ for the KFTT Japanese-English experiments and $\alpha=1.0$ for the rest. We evaluate our models using the BLEU \cite{Papineni:2002:BMA:1073083.1073135} metric implemented in tensor2tensor as {\it t2t\_bleu}. In order to generate pseudo-parallel corpora by back-translating the GCP monolingual corpora, we used the 1-layer NMT models for decoding. In order to save time we perform greedy decoding\footnote{We could translate approximately 1.5 million lines in approximately 40 minutes using 8 GPUs.}. For the GCP English-Japanese translation direction, we also tried to see what happens if a model is trained using N-layers of recurrence but is decoded using fewer than N-1 layers and more than N layers of recurrence.

\begin{table*}[t]
\centering
\begin{tabular}{|c|c|c|c|c|c|c|}
\hline
\multirow{3}{*}{\textbf{\begin{tabular}[c]{@{}c@{}}\#recurrently\\ stacked \\ layers\end{tabular}}} & \multicolumn{6}{c|}{\textbf{BLEU}}                                                                      \\ \cline{2-7} 
                                                                                                  & \multicolumn{2}{c|}{\textbf{WMT}} & \textbf{ALT}   & \textbf{KFTT}  & \multicolumn{2}{c|}{\textbf{GCP}} \\ \cline{2-7} 
                                                                                                  & \textbf{Tr-En}  & \textbf{En-Tr}  & \textbf{Ja-En} & \textbf{Ja-En} & \textbf{Ja-En}  & \textbf{En-Ja}  \\ \hline
\textbf{1}                                                                                        & 13.75           & 11.94           & 7.59           & 21.64          & 21.95           & 23.89           \\ \hline
\textbf{2}                                                                                        & 15.95           & 14.62           & 7.60            & 24.50           & 23.24           & 24.47           \\ \hline
\textbf{3}                                                                                        & 16.39           & 14.68           & 7.99           & 25.84          & 23.42           & 25.02           \\ \hline
\textbf{4}                                                                                        & 17.05           & 14.93           & 7.91           & 26.23          & 24.33           & 25.28           \\ \hline
\textbf{5}                                                                                        & 17.12           & 15.51           & \textbf{8.28}  & 26.42          & 23.95           & 25.38           \\ \hline
\textbf{6}                                                                                        & \textbf{17.31}  & \textbf{15.77}  & 8.26           & \textbf{26.51} & \textbf{24.36}  & \textbf{25.84}  \\ \hline
\textbf{6-layer model}                                                                            & 18.36           & 16.29           & 8.47           & 27.19          & 24.67           & 26.22           \\ \hline
\end{tabular}
\caption{Results for our experiments on WMT, ALT, KFTT and the GCP data-sets. The last row denoted as `6-layer model` is a transformer model with 6-layers without any parameter sharing. The results in bold indicate the best values among the diverse models of the proposed method.}
\label{allresults}
\end{table*}

\begin{table*}[t]
\centering
\begin{tabular}{|c|c|c|c|c|c|c|}
\hline
\multirow{2}{*}{\textbf{\begin{tabular}[c]{@{}c@{}}\#recurrence\\ for decoding\end{tabular}}} & \multicolumn{6}{c|}{\textbf{\#recurrently stacked layers in model}}                                 \\ \cline{2-7} 
                                                                                              & \textbf{1}     & \textbf{2}     & \textbf{3}     & \textbf{4}     & \textbf{5}     & \textbf{6}     \\ \hline
\textbf{1}                                                                                    & \textbf{23.89} & 19.57          & 11.32          & 6.16           & 4.43           & 2.56           \\ \hline
\textbf{2}                                                                                    & 21.05          & \textbf{24.47} & 23.63          & 20.72          & 15.53          & 11.10           \\ \hline
\textbf{3}                                                                                    & 12.33          & 23.91          & \textbf{25.02} & 25.03          & 23.67          & 20.61          \\ \hline
\textbf{4}                                                                                    & 5.06           & 22.06          & 24.89          & \textbf{25.28} & 25.18          & 24.82          \\ \hline
\textbf{5}                                                                                    & 2.13           & 19.84          & 23.95          & 24.78          & \textbf{25.38} & 25.81          \\ \hline
\textbf{6}                                                                                    & 0.88           & 17.45          & 22.63          & 24.09          & 24.75          & \textbf{25.84} \\ \hline
\textbf{7}                                                                                    & 0.40            & 15.34          & 21.45          & 23.26          & 24.56          & 25.53          \\ \hline
\textbf{8}                                                                                    & 0.21           & 12.47          & 20.03          & 22.36          & 24.11          & 24.87          \\ \hline
\end{tabular}
\caption{Results of decoding the GCP English-Japanese model for different levels of recurrence stacking than what was used for training.}
\label{recurrentdecoding}
\end{table*}

\begin{table}[]
\centering
\begin{tabular}{|c|c|c|c|c|}
\hline
\multirow{4}{*}{\textbf{\begin{tabular}[c]{@{}c@{}}\#recurrently\\ stacked\\ layers\end{tabular}}} & \multicolumn{4}{c|}{\textbf{Language Direction}}                          \\ \cline{2-5} 
                                                                                                 & \multicolumn{4}{c|}{\textbf{Back-translated Data Used}}                    \\ \cline{2-5} 
                                                                                                 & \multicolumn{2}{c|}{\textbf{Ja-En}} & \multicolumn{2}{c|}{\textbf{En-Ja}} \\ \cline{2-5} 
                                                                                                 & \textbf{No}     & \textbf{Yes}      & \textbf{No}     & \textbf{Yes}      \\ \hline
\textbf{1}                                                                                       & 21.95           & \textbf{23.90}     & 23.89           & \textbf{25.47}    \\ \hline
\textbf{2}                                                                                       & 23.24           & \textbf{24.79}    & 24.47           & \textbf{26.56}    \\ \hline
\textbf{3}                                                                                       & 23.42           & \textbf{24.79}    & 25.02           & \textbf{26.66}    \\ \hline
\textbf{4}                                                                                       & 24.33           & \textbf{25.17}    & 25.28           & \textbf{27.31}    \\ \hline
\textbf{5}                                                                                       & 23.95           & \textbf{24.92}    & 25.38           & \textbf{27.08}    \\ \hline
\textbf{6}                                                                                       & 24.36           & \textbf{25.82}    & 25.84           & \textbf{27.55}    \\ \hline
\end{tabular}
\caption{Using back-translated data for GCP corpus setting. The translation quality increases despite the number of parameters remaining the same.}
\label{backtransresults}
\end{table}
\section{Results and Discussion}
\subsection{Main Results}
Refer to Table~\ref{allresults} for the results of the experiments using up to 6-layers of recurrently stacked layers for the WMT, ALT, KFTT and GCP data-sets. We observed that, no matter the data-set, the translation quality improves as the same parameters are recurrently used in a depth-wise fashion. The most surprising result is that the performance of our 6-layer recurrently stacked model with shared parameters across all layers approaches the performance of the vanilla 6-layer model without any parameter sharing across layers. The most probable explanation of this phenomenon is that the parameter sharing forces the higher layers of the NMT model to learn more complex features. Note that dropout is applied by default in the transformer implementation we use. Thus, at each stage, the same set of parameters has to make do with less reliable representations. This means that the representations at the topmost layers are very robust and thus enable better translation quality. The gains in translation quality are inversely proportional to the amount of recurrent stacking layers. 

In the case of the ALT corpus, the performance trends are not very clear. Although, there are improvements in translation quality with each level of recurrent stacking, they are not very sharp as observed for the other language directions. We suspect that this is because of the extremely low resource setting in which NMT training is highly unreliable. Nevertheless, we do not see any detrimental effects of recurrent stacking of layers.

\subsection{Decoding Using Different Recurrence Steps}
In order to understand what each step of recurrent stacking brings about, we decided to train a N-layer recurrently stacked model and during decoding perform recurrence up to N-1 times. Refer to Table~\ref{recurrentdecoding} for the results of the same on the GCP English-Japanese translation. It can be seen that once the NMT model has been trained to use N layers of recurrent stacking, it is unable to perform optimally using fewer than N levels of stacking for decoding. Although, this is expected, there are three crucial observations as below.

Firstly, the computation of the most useful and hence the most complex features takes place at higher levels of recurrence. For a 6-layer recurrent stacking model, using just 1-layer (no recurrence) during decoding, gives a BLEU of 2.56. However, as we perform more layers of recurrence, the BLEU jumps drastically. This could imply that the NMT model delays the learning of extremely complex features till the very end. Secondly, for the same model, the difference between using the full 6-layer recurrent stacking and 5-layer recurrent stacking is not very significant. This means that as we train the model using a large number of recurrent stacking, it is possible to use fewer layers of recurrent stacking for decoding. Thirdly, when we use the more than 6-layer recurrent stacking, the BLEU score starts dropping again (25.53 and 24.87 for 7 and 8 layers during decoding). This indicates that the model has not learned to extract complex features beyond what it has been trained for. However, once the model has been trained for non-zero number of recurrences, the drop in quality is less severe as can be seen for a 3-layer recurrent model being decoded for more than 3 layers of recurrence. In the future, we plan to see what happens when we train a model using beyond 6 layers of recurrent stacking during training and decoding to identify the limits of recurrence.

\subsection{Using Back-translated Corpora}
As mentioned in the experimental section, for the GCP corpus setting, we generated pseudo-parallel corpora by translating monolingual corpora of 1552475 lines using the 1-layer models\footnote{To translate the Japanese monolingual corpus we used the 1-layer Japanese-English model.}. We added this pseudo-parallel corpora to the original parallel corpora of 400000 lines and trained all models mentioned in Section~\ref{sec:nmtmodels} from scratch. In order to compensate for the additional data we trained, both, the Japanese-English and the English-Japanese models for 200000 iterations on 1 GPU. Table~\ref{backtransresults} provides the results for models up to 6-layer recurrent stacking.

We can see that despite no increase in the number of parameters, the presence of back-translated data augments the translation quality for both translation directions. For the English-Japanese translation, the quality of a 3-layer recurrently stacked model trained using additional back-translated data matches the quality of a vanilla 6-layer model trained on the original parallel corpus of 400000 lines. Furthermore, the 6-layer recurrently stacked model beats the 6-layer model trained on the original parallel corpus of 400000 lines. It is clear that the gains using additional layers of recurrence in a low resource scenario is much higher than the gains in a resource rich scenario.

\subsection{Parameter Reduction Due to Sharing}
The number of parameters in a vanilla 6-layer Turkish-English model is 158894599 whereas the number of parameters for the recurrent stacking layer models is 48640519 no matter how many layers are in the stack. This corresponds to a 3.26 times reduction in the number of parameters. Knowledge distillation could help reduce this size even further. Similarly, for English-Japanese, the recurrently stacked models are 2.12 times smaller than the vanilla 6-layer model\footnote{Note that the Turkish-English models have a shared matrix for encoder embedding, decoder embedding and softmax and thus will have greater savings in terms of parameters}.


\section{Conclusion}
In this paper, we have proposed a novel modification to the NMT architecture where we share parameters across the layers of a N-layer model leading to a recurrently stacked NMT model. As a result, our model has the same size as that of a single layer NMT model and gives comparable performance to a 6-layer NMT model where the parameters across layers are not shared. This shows that it is possible to train compact NMT models without a significant loss in translation quality. We also showed that our approach can be used to generate pseudo-parallel corpora or back-translated corpora which when added to the original parallel corpora leads to further improvements in translation quality. We believe that our work will promote the research of techniques that rely on reusability of parameters and hence simplify the existing NMT architectures. In the future, we will perform an in-depth analysis of the limits of recurrent stacking of layers in addition to combining our methods with knowledge distillation approaches for high performance compact NMT modeling. We also plan to experiment with more complex mechanisms to compute the recurrent information during stacking for improved NMT performance.

\bibliography{acl2018}
\bibliographystyle{acl_natbib}

\end{document}